\documentclass{article}
\usepackage{arxiv}

\usepackage{cite}
\usepackage{amsmath,amssymb,amsfonts}
\usepackage{algorithmic}
\usepackage
{graphicx}
\usepackage{textcomp}
\usepackage{xcolor}
\def\BibTeX{{\rm B\kern-.05em{\sc i\kern-.025em b}\kern-.08em
    T\kern-.1667em\lower.7ex\hbox{E}\kern-.125emX}}

\usepackage[english]{babel}
\usepackage{bm}
\usepackage{url}

\usepackage{amsmath}
\usepackage{graphicx,dsfont}
\usepackage[colorlinks=true, allcolors=blue]{hyperref}

\usepackage{graphicx}
\usepackage[utf8]{inputenc}

\usepackage{amsmath, amsxtra, amsfonts, amssymb, amstext, mathtools}
\usepackage{booktabs}
\usepackage{makecell}
\usepackage{multicol}
\usepackage{multirow}
\usepackage{longtable}
\usepackage{array,xspace}

\usepackage{rotating}

\newcommand{\ignore}[1]{}

\usepackage{color}

\usepackage{xcolor, soul}
\sethlcolor{yellow}
\newcommand{\cmmnt}[1]{}
\usepackage{algorithm}
\usepackage{algorithmic}
\usepackage{amsmath}
\usepackage{multirow}
 \usepackage{float}
\usepackage{adjustbox}

\graphicspath{ {./} }

\begin{document}

\title{Evolving Reliable Differentiating Constraints for the Chance-constrained Maximum Coverage Problem}

\author{Saba Sadeghi Ahouei\\
Optimisation and Logistics\\
School of Computer and Mathematical Sciences\\
The University of Adelaide\\
Adelaide, Australia
\And
Jacob de Nobel\\
Leiden Institute of Advanced Computer Science\\
Leiden University\\
Leiden, The Netherlands
\And
Aneta Neumann\\
Optimisation and Logistics\\
School of Computer and Mathematical Sciences\\
The University of Adelaide\\
Adelaide, Australia
\And
Thomas B{\"a}ck\\
Leiden Institute of Advanced Computer Science\\
Leiden University\\
Leiden, The Netherlands
\And
Frank Neumann\\
Optimisation and Logistics\\
School of Computer and Mathematical Sciences\\
The University of Adelaide\\
Adelaide, Australia
}

\maketitle
\begin{abstract}
Chance-constrained problems involve stochastic components in the constraints which can be violated with a small probability. We investigate the impact of different types of chance constraints on the performance of iterative search algorithms and study the classical maximum coverage problem in graphs with chance constraints. Our goal is to evolve reliable chance constraint settings for a given graph where the performance of algorithms differs significantly not just in expectation but with high confidence. This allows to better learn and understand how different types of algorithms can deal with different types of constraint settings and supports automatic algorithm selection. We develop an evolutionary algorithm that provides sets of chance constraints that differentiate the performance of two stochastic search algorithms with high confidence. We initially use traditional approximation ratio as the fitness function of (1+1)~EA to evolve instances, which shows inadequacy to generate reliable instances. To address this issue, we introduce a new measure to calculate the performance difference for two algorithms, which considers variances of performance ratios.
Our experiments show that our approach is highly successful in solving the instability issue of the performance ratios and leads to evolving reliable sets of chance constraints with significantly different performance for various types of algorithms.


\end{abstract}

\section{Introduction}
It is not possible for one algorithm to perform perfectly on all optimization problems \cite{DBLP:journals/tec/DolpertM97}. Some algorithms work better on a range of problems with specific characteristics than others. It is important to investigate the performance and behavior of different algorithms on a wide range problems in order to be able to select the best algorithms to use for specific types of problems. 
This problem which is known as the algorithm selection problem~\cite{DBLP:journals/ac/Rice76} is considered as a learning problem in the machine learning community. The goal is to learn the relation between the performance of algorithms and characteristics of problems. Instances with different complexity levels, a wide range of algorithms and suitable performance metrics are needed for solving algorithm selection problem as a meta-learning problem. Such materials combined create a comprehensive set of meta-data about the algorithms' performance that can be used for automated algorithm selection, and classification of algorithms based on problems’ features. ~\cite{DBLP:journals/csur/Smith-Miles08,DBLP:journals/ec/KerschkeHNT19,DBLP:conf/cec/NeumannP16,DBLP:journals/ec/GaoNN21}.

As mentioned above, we need a set of benchmark problems with varying degrees of complexity to be able to learn which algorithms are suitable for which kind of problems.  Benchmark problems that can accentuate the strengths and weaknesses of an algorithm in solving a specific type of problem are very beneficial when observing the behavior of algorithms. Discriminating instances are capable to do this by increasing the difference in performance between two selected algorithms. 
In this paper, we aim to evolve chance-constrained maximum coverage problem instances that are easy to solve for one algorithm and hard to solve for another with high confidence.
Maximum coverage is a monotone submodular problem. Submodular functions can model a wide range of real-world problems \cite{DBLP:conf/gecco/NeumannB021,DBLP:journals/mp/NemhauserWF78,DBLP:journals/combinatorica/Cunningham85}. 
In submodular problems, we have diminishing returns, which means adding more elements to the solution decreases the addition of the elements’ benefit \cite{DBLP:conf/aaai/DoerrD0NS20,DBLP:conf/ijcai/DoGN023}. 
Optimization of deterministic submodular functions has been widely studied \cite{DBLP:journals/mp/NemhauserWF78, DBLP:journals/mor/NemhauserW78, DBLP:journals/siamcomp/CalinescuCPV11, DBLP:journals/siamcomp/FeigeMV11, DBLP:conf/soda/BuchbinderFNS14,DBLP:conf/ecai/Doskoc00N0Q20,DBLP:conf/ppsn/FriedrichN14,DBLP:journals/ec/FriedrichN15,DBLP:conf/ppsn/DoN20}. However, chance-constrained submodular optimization has just gained attention recently \cite{DBLP:conf/ecai/YanD0Q023}. Chance-constrained optimization is one of the methods to address uncertainty. Many real-world problems contain uncertainty, thus, considering uncertainty can facilitate us to have problems that are closer to real-world applications. Lack of reliability is one of the issues that arises when it comes to uncertainty. In this study, we address this issue while evolving differentiating instances for the chance-constrained maximum coverage problem. We introduce a new approach to evolve instances that are easy to solve for one and hard to solve for another algorithm in the chance-constrained setting, which leads to generating instances with higher confidence and lower variance despite the method that has been used by previous studies.

\subsection{Related work}
Uncertainty is a crucial part of optimization problems in practice which is addressed by different approaches. Chance-constrained optimization is one of the techniques used to tackle uncertainty. A chance constraint involves random components and can be violated with a small probability. Evolutionary algorithms have been successfully applied to chance constrained optimization problems \cite{DBLP:journals/eor/PoojariV08,DBLP:conf/gecco/XieN020,E23OMOEA,U3OEADCHKP,TCHTTP,POCCMSP,DBLP:conf/ppsn/NeumannN20,DBLP:journals/tec/LiuZFG13}. Xie et al. \cite{DBLP:conf/gecco/XieHAN019} developed (1+1)~EA and multi-objective evolutionary algorithm (GSEMO) to solve the chance-constrained knapsack problem. They employed Chebyshev's inequality and Chernoff bound to evaluate constraint violations. Their experiments indicated that GSEMO provides better solutions for this problem. 
Doerr et al. \cite{DBLP:conf/aaai/DoerrD0NS20} utilized a greedy algorithm to optimize chanced-constrained submodular functions. They solved the chance-constrained influence maximization problem with the greedy algorithm and compared the results with the deterministic problem. Based on their investigations, they found that the greedy algorithm was able to obtain high-quality solutions for this problem. Neumann and Neumann \cite{DBLP:conf/ppsn/NeumannN20} proposed a global simple evolutionary multi-objective optimizer (GSEMO) to solve monotone submodular functions. To evaluate the probability of violating the chance constraint, they used Chebyshev's inequality and Chernoff bound. The fitness function in this algorithm consists of two objectives, one ensures the feasibility of solutions and the other handles their optimality. They also conducted a runtime analysis of GSEMO and performed experiments on the maximum influence problem and the maximum coverage problem with uniform weights. Experimental results indicated that GSEMO outperforms the greedy algorithm for such problems. Xie et al. \cite{DBLP:conf/gecco/XieN0S21} investigated the runtime of the randomized local search algorithm (RLS) and (1+1)~EA for the chance-constrained knapsack problem in a new setting, where components have correlated uniform weights. They proved that RLS and (1+1)~EA can find feasible solutions within an expected time of less than $O(n \log n)$ and $O(n^2 \log n)$, respectively. Moreover, they introduced two additional settings, in the first setting components have uniform profits, while in the second one, components are divided into several groups and each group has the same arbitrary profit. The authors analyzed how these profit settings affected the behavior and runtime bounds of the two above-mentioned algorithms. 
Neumann and Witt \cite{DBLP:conf/ijcai/0001W22} provided an analysis of the solutions generated by (1+1)~EA for chance-constrained combinatorial optimization problems with normally distributed components, subject to spanning tree constraints. They showed that (1+1)~EA does not work efficiently in this setting and tends to converge to a local optimum rather than the global optimum. To address this limitation, they developed a multi-objective approach with a trade-off between expected cost and variance, resulting in better solutions for the problem. Finally, they further improved this algorithm by introducing the convex GSEMO.
Neumann et al. \cite{DBLP:conf/ppsn/NeumannXN22} considered a chance-constrained knapsack problem with stochastic profits and deterministic weights, dynamic constraints \cite{MOEAwSWS} in which a level of profit is guaranteed with a high level of confidence. They used Chebyshev's inequality and Hoeffding Bound to handle the chance constraint and compared the performance of three different algorithms in solving this problem. Shi et al. \cite{DBLP:conf/ppsn/ShiYN22} studied computational complexity and runtime analysis of (1+1)~EA and RLS for two different versions of chance-constrained makespan scheduling problem.

Benchmarking is a systematic analysis of the performance of one or several algorithms on one or a class of optimization problems \cite{DBLP:journals/asc/DoerrYHWSB20}. Benchmarking has a wide range of applications in optimization, such as comparing the performance of different algorithms on a set of problems, assessing the performance of a new algorithm for a particular type of problem, algorithm selection and, configuration. Furthermore, benchmarking is beneficial for identifying gaps and issues in theoretical studies \cite{DBLP:journals/corr/abs-2007-03488}. To have reliable studies in this area, having a wide variety of benchmark instances is crucial. In algorithm selection, instances that can show significant differences in algorithms' performance, help with finding the strengths and weaknesses of algorithms in solving a specific problem.
There are some studies around generating benchmark instances with significant performance differences between two algorithms. Gao et al. \cite{DBLP:journals/ec/GaoNN21} introduced a new evolutionary algorithm to generate diverse easy/hard instances for the traveling salesman problem considering different features of the problem. This approach resulted in more coverage over feature space. They proceed to classify those instances using the support vector machines (SVM) classification model. Bossek et al. \cite{DBLP:conf/foga/BossekKN00T19} introduced new mutation operators to evolve instances for the traveling salesman problem. Using these new mutation operators led to more diverse instances even without employing diversity strategies. In the body of literature for chance-constrained optimization, benchmarking is rarely explored.
Neumann et al. \cite{DBLP:conf/cec/NeumannNQDNVAYWB23} 
provided new instances for chance-constrained submodular problems. They proceed to test the performance of 12 baseline algorithms on those instances using the IOHprofiler framework. One of the monotone chance-constrained problems that they investigate is chance-constrained maximum coverage problems. Their experiments show that there is no noticeable difference between the performance of these algorithms in solving these benchmark instances. As stated previously, it is essential to have reliable discriminating instances to be able to study how different algorithms perform in solving a particular problem. 

\subsection{Our Contribution}
To the best of our knowledge, there is no set of instances that shows the difference between baseline algorithms' performance in solving chance-constrained submodular problems. In this paper, we introduce an evolutionary algorithm that can evolve sets of reliable instances for the chance-constrained maximum coverage problem, that are easy to solve for one algorithm and hard to solve for the other. We utilize a (1+1)~EA to generate discriminating instances for different instance sizes of chance-constrained maximum coverage problems. 

Initially, we use the traditional method to calculate the performance difference. The instances generated by this method demonstrate the lack of reliability in the performance ratio in the chance-constrained setting due to the uncertainty involved in the problem as well as the performance of randomized algorithms among their runs. We cannot base any future studies on unreliable benchmark instances. Hence, it is important to implement a method that tackles this issue. In order to get instances with a higher confidence level, we formulate a new performance measure as the fitness function of our EA, which takes into account the variances of the performance ratios.
Our goal is to increase the confidence level of our results while getting a high difference in the performance of two selected algorithms.

We increase the performance difference between the two selected algorithms by maximizing this ratio using (1+1)~EA. The fitness function that we have used demonstrates a great ability to decrease the standard deviation of the performance ratio and generate much more stable instances. These instances help one to understand baseline algorithms' advantages and limitations in solving the chance-constrained maximum coverage problem.

The paper is structured as follows.
In Section~\ref{sec2}, we introduce the chance-constrained maximum coverage problem as a monotone submodular optimization problem, in Section~\ref{sec3} we proposed a (1+1)~EA to evolve sets of constraints for this problem which makes it hard to solve by one algorithm and easy to solve by the other. We utilized the approximation ratio as the fitness function of our evolutionary algorithm in this section. In Section~\ref{sec4} we introduce the discounting approximation ratio as a new technique to measure the performance difference. We use this new fitness function to evolve instances with a high level of confidence.
Eventually, in Section~\ref{sec5} we represent experimental results using these two approaches to evolve discriminating instances and show the new discounting fitness function that developed gives us more reliable instances in compared to the approximation ratio, which is widely used in studies before.

\section{Preliminaries}
\label{sec2}

In this paper, we aim to evolve discriminating instances for the chance-constrained maximum coverage problem with high confidence. Chance-constrained maximum coverage is a stochastic monotone submodular problem. Given an undirected weighted graph $G=(V,\;E,\;c)$, this problem aims to choose a subset of vertices $V'$, in a way that they would cover the maximum number of the nodes in the graph at least by one endpoint, while the cost can exceed the budget $B$ just by a small probability $\alpha$.


We consider a given search point $x\in\{0,1\}^n$ where $n=|V|$, $V'(x)=\{v_i|x_i=1\}$, and $N(V'(x))$ contains the vertices in  $V'$ and all of their neighbours. The chance-constrained maximum coverage problem can be formulated as follows:
\begin{align}
        \text{Maximize} \qquad f(x)  =N(V'(x)) \\
        \text{Subject to} \qquad    Pr(C(x)>B)\leq\alpha
        \label{eq:C}
\end{align}

In the stochastic setting, a non-negative uniformly at random cost $c_v$ with expected value $E(c_v)=\mu_v ,\; \mu_v\in[0,\;\mu_{max}]$ and variance $Var(c_v)=\sigma^2_v,\;\sigma^2_v\in[0,\;\mu^2_v/3]$ is assigned to each node.
The cost of the nodes are independent random variables. Hence, for a given solution $X=(x_1, x_2,...,x_n)\in \{0,1\}^n$ we have:
\begin{displaymath}
C(X)= \sum_{i=1}^{n} C_i \cdot x_i
\end{displaymath}
\noindent with expected value and variance:
\begin{eqnarray*}
    E[C(X)] & = \sum_{i=1}^{n} \mu_i \cdot x_i \\
    Var[C(X)] & = \sum_{i=1}^{n} \sigma^2_i \cdot x_i
\end{eqnarray*}

We evolve the new instances by modifying these expected values and variances, without changing the initial graphs. To handle the chance constraint we introduce a surrogate function that approximately calculates the probability of constraint violation, using one-sided Chebyshev's Inequality. According to \cite{DBLP:conf/gecco/XieHAN019}, if the expected weight and variance of the solution $X$ are known, an upper bound for the chance constraint (Equation~\ref{eq:C}) is calculated as follows utilizing Chebyshev's inequality:
\begin{equation} \label{eq:C2}
     Pr(C(X)>B)\leq \frac{Var[C(X)]}{Var[C(X)]+(B-E[C(X)])^2}
\end{equation}

Therefore, for any $\alpha\geq\frac{Var[C(X)]}{Var[C(X)]+(B-E[C(X)])^2}$, solutions that are feasible according to Equation~\ref{eq:C2}, also will satisfy the chance constraint (Equation~\ref{eq:C}). According to Equation~\ref{eq:C2}, and considering expected values and variances of solutions in our setting, for each solution the surrogate function $\beta$ is defined as:
\begin{equation}
    \beta (X) = \frac {\sum_{i=1}^{n} \sigma^2_i \cdot x_i}{\sum_{i=1}^{n} \sigma^2_i \cdot x_i + (B-\sum_{i=1}^{n} \mu_i \cdot x_i)}
\end{equation}

This surrogate function is used as an upper bound for the capacity constraint in our problem and as mentioned before any solution that is feasible according to this upper bound, is considered as a feasible solution for the problem.

\section{Evolving Discriminating Instances Using Approximation Ratio} \label{sec3}

To generate instances that are hard to solve for one algorithm and easy to solve for the other, it is necessary to have a method that measures the difference in performance between the two selected algorithms for that particular problem. In previous studies, the most widely-used performance measure is defined as follows \cite{DBLP:journals/ec/GaoNN21}:
\begin{equation} \label{eq:FE1}
f(X)=P(A_1)/P(A_2)
\end{equation}

\noindent where $P(A_1)$ and $P(A_2)$ are the performance of the first and the second algorithm, respectively. Our study demonstrates this performance measure is too noisy and not reliable for chance-constrained problems. We refer to this as the approximation ratio or traditional method in the following.

\begin{algorithm}[t]
\caption{$(1+1)~EA$}\label{alg:main}
\begin{algorithmic}

\STATE Generate an instance with uniformly at random expected values $E(X)$ and variances $V(X)$.
\WHILE {$stopping\: criteria\: is\: not\: met$}
\FOR{$each\: node$}
\STATE with probability $P_m$
\STATE choose "a" randomly from $N(0,\sigma_1)$
\STATE choose "b" randomly from $N(0,\sigma_2)$
\STATE $E(Y) \gets \min(\mu_{max},\max(0,E(X)+a)$
\STATE $Var(Y) \gets \min(Var_{max},\max(0,Var(X)+b)$
\ENDFOR
\STATE  update the value of budget B(Y) with new expected values.
\IF {f(X) $\geq$ f(Y)}
    \STATE X $\gets$ Y
    
\ENDIF
\ENDWHILE
\end{algorithmic}

\end{algorithm}

\subsection{Evolutionary Approach}

Many of the previous studies about evolving instances use deterministic costs \cite{DBLP:conf/foga/BossekKN00T19, DBLP:journals/ec/GaoNN21,DBLP:journals/amai/MersmannBT0BN13}. However, we consider a problem that involves uncertainty which is closer to real-world problems. This chance-constrained maximum coverage problem consists of deterministic profits and non-deterministic costs, associated with each node of a graph. We consider these costs as uniformly at random uncorrelated values. In this study, we aim to generate differentiating sets of constraints by changing the value of expected values and variances associated with the cost of each node by using an evolutionary algorithm. The (1+1)~EA initializes with an individual with non-negative uniformly at random costs. These stochastic costs are represented as $n$ independent uniformly at random variables each with expected values $(0,\;\mu_{max}]$, and variances $(0,\;\mu_i^2/3]$, which allocates to each node $i$, a uniform cost with a maximum range of $2\cdot\mu_i$.
Then the constraint budget will be calculated according to these weights using 
$B=n/30 \cdot (\mu_{max})/2$, our experiments indicate that this budget provides us with flexibility in adjusting costs, without being excessively large. This prevents the problem from becoming too easy, as a large budget could negate the effect of changing the costs on the problem difficulty.
These expected values and variances will be changed in mutation in order to create offspring for the next generation of the EA. We use an elitist strategy to select the best individual in every iteration. This selection function continuously keeps the individual with a larger performance ratio between the parent and offspring, in every iteration. This expedites convergence towards optimal or near-optimal solutions. 

By maximizing the performance difference between the selected pair of algorithms throughout the optimization process, we would be able to evolve instances that are easy to solve for $A_1$ and hard to solve for $A_2$ with high confidence.

\subsection{Mutation Operator}
As we previously stated, for every instance, there is an expected value and variance associated with the cost of each vertex. For the mutation process, we introduce an operator that changes the expected value and variance of each node independently with the probability of $Pm$. In the mutation process, random values "$a$" and "$b$" are chosen from normal distributions $N(0,\sigma_w)$ and $N(0,\sigma_v)$ independently, and are added to the expected value and variance of each node. If the new value is out of range we set it to the value of the bounds. $\sigma_w$ and $\sigma_v$ are adapting automatically, using 1/5 success rule, throughout the optimization process. Eventually, the budget is updated considering the new expected values. In this mutation function, by applying small modifications to a large number of nodes in each iteration, we make sure the algorithm explores the solution space effectively.

\section{Evolving Reliable Differentiating Chance Constraints} \label{sec4}
The performance of randomized algorithms such as evolutionary algorithms varies among their runs. This implies that generated instances might not be reliable in terms of differentiating the performance of algorithms.
We now introduce our new evolutionary approach for evolving reliable differentiating instances. Our goal is to maximize the performance ratio based on several runs of the considered algorithms. To create instances that differentiate the performance in a reliable way we take into account the standard deviation across the different runs.

\subsection{Fitness function taking into account reliability}

In order to get more stable instances we define discounting expected approximation ratio based on uncertainty (Equation \ref{eq:FE2}). This new performance measure takes into account variances of these performance ratios obtained in each run, which limits the effect of uncertainty in the fitness function and helps to get more stable solutions.
To evaluate the performance of each algorithm we calculate the expected value and standard deviation of their function values (best value that it reaches for that instance) for several independent runs. If the best value that an algorithm can reach in any of these runs is negative, which means it is not a feasible solution, we replace that negative value with a small value $\epsilon$. For each run $i$, $f(x_i)=P(A_1i)/P(A_2i)$ and the performance ratio is formulated as:
\begin{equation}\label{eq:FE2}
    f^\prime(X)= E[f(x)] - k_\alpha \cdot  std[f(X)]
\end{equation}

\noindent where $ E[f(x)]$ and $std[f(X)]$ are the expected value and standard deviation of performance ratios in $n$ independent runs, 
and $k_\alpha$ represents a constant determined by the desired confidence level $\alpha$. By adjusting the value of $k_\alpha$ we can make a trade-off between expected performance ratios and the variance of these ratios. By increasing the values of $k_\alpha$ we adopt a more conservative approach that favors solutions with lower standard deviation to the ones with higher expected performance ratios \cite{DBLP:conf/cec/StimsonRNRN23}. This formula has been used in previous studies on chance-constrained problems to enhance the confidence level in solutions by decreasing the effect of uncertainty \cite{DBLP:conf/ppsn/NeumannXN22, DBLP:conf/cec/StimsonRNRN23, DBLP:conf/gecco/0001W23}. However, to best of our knowledge it has not been used for evolving instances yet.

Given a pair of algorithms, we aim to generate reliable instances that are easy to solve for one algorithm and hard to solve for the other. To achieve that, we developed a (1+1)~EA (demonstrated in Algorithm~\ref{alg:main}) which evolves these instances by changing the expected values and variances of uniform costs of each vertex.

\section{Experimental Investigations} \label{sec5}

In this section, we investigate the experimental results for evolving reliable sets of instances with high performance differences. The proposed method in this paper can be used for a wide range of optimisation problems and pairs of algorithms. For our experimental investigations, we are using chance-constrained maximum coverage problem instances which are implemented in the IOHProfiler framework. IOHProfiler is a benchmarking tool that helps researchers design, compare, and analyze iterative optimization heuristic algorithms \cite{DBLP:journals/corr/abs-1810-05281}. It consists of two core components. One module is IOHexperimenter, which is a flexible platform that researchers can use to create new problems and algorithms or utilize predefined ones for benchmarking \cite{DBLP:journals/corr/abs-2111-04077}. The other one is IOHAnalyser, which is an open-source interactive platform that provides visual representations of benchmarking data to help with a better understanding of the behavior of the algorithms on different problems throughout the optimization process \cite{DBLP:conf/gecco/WangVYDB22}. We generate instances that are easy for (1+1)~EA and hard for FGA and GHC in our experiments. These algorithms are already implemented in the IOHProfiler. 
Details of these algorithms can be found in 
Doerr et al. \cite{DBLP:journals/asc/DoerrYHWSB20}.
\\ We use (1+1)~EA with two different fitness functions to evolve these instances. First, we use the performance ratio as the fitness function. This measure is the ratio of the best objective value reached by the two selected algorithms and shows the performance of those algorithms in comparison. This fitness function does not have the capability to generate reliable and stable instances when uncertainty is involved in the components. To address this issue, we then implement the discounting method which takes into account the standard deviation of performance ratios. Using this method eliminates solutions with high variances throughout the optimization process and results in instances with higher confidence.

\begin{table}
\centering
 \caption{Values of the two mentioned function values for 1000 random instances, each algorithm run for 10 separate runs and 10,000 function evaluations}\label{tab:random10}
\renewcommand{\tabcolsep}{7pt}
\renewcommand{\arraystretch}{1.1}
\begin{tabular}{@{}cccccccc@{}}
\toprule
\multirow{3}{*}{\begin{tabular}[c]{@{}c@{}}fitness\\  function\end{tabular}} & \multicolumn{7}{c}{algorithm budget = 10,000} \\ \cmidrule(l){2-8} 
 & \multicolumn{7}{c}{performance ratio} \\ \cmidrule(l){2-8} 
 & \begin{tabular}[c]{@{}c@{}}easy\\ hard\end{tabular} & \begin{tabular}[c]{@{}c@{}}EA\\ SA\end{tabular} & \begin{tabular}[c]{@{}c@{}}SA\\ EA\end{tabular} & \begin{tabular}[c]{@{}c@{}}EA\\ FGA\end{tabular} & \begin{tabular}[c]{@{}c@{}}FGA\\ EA\end{tabular} & \begin{tabular}[c]{@{}c@{}}SA\\ FGA\end{tabular} & \begin{tabular}[c]{@{}c@{}}FGA\\ SA\end{tabular} \\ \midrule
\multirow{4}{*}{\begin{tabular}[c]{@{}c@{}}performance \\ ratio\end{tabular}} & average & 0.984 & 1.019 & 1.081 & 0.930 & 1.100 & 0.914 \\
 & std & 0.018 & 0.019 & 0.029 & 0.024 & 0.033 & 0.026 \\
 & min & 0.905 & 0.972 & 1.002 & 0.855 & 1.017 & 0.827 \\
 & max & 1.032 & 1.107 & 1.185 & 1.001 & 1.221 & 0.984 \\ \midrule
\multirow{4}{*}{\begin{tabular}[c]{@{}c@{}}discounting\\ expected\\ approximation\\ ratio\end{tabular}} & average & 0.874 & 0.904 & 0.913 & 0.788 & 0.934 & 0.779 \\
 & std & 0.039 & 0.035 & 0.046 & 0.051 & 0.043 & 0.052 \\
 & min & 0.705 & 0.781 & 0.693 & 0.606 & 0.758 & 0.559 \\
 & max & 0.974 & 0.995 & 1.060 & 0.921 & 1.047 & 0.898 \\ \bottomrule
\end{tabular}
\end{table}

\begin{table}
\centering
 \caption{Values of the two mentioned function values for 1000 random instances, each algorithm run for 10 separate runs and 100,000 function evaluations 
 }\label{tab:random100}
\renewcommand{\tabcolsep}{7pt}
\renewcommand{\arraystretch}{1.1}
\begin{tabular}{@{}cccccccc@{}}
\toprule
\multirow{3}{*}{\begin{tabular}[c]{@{}c@{}}fitness\\  function\end{tabular}} & \multicolumn{7}{c}{algorithm budget = 100,000} \\ \cmidrule(l){2-8} 
 & \multicolumn{7}{c}{performance ratio} \\ \cmidrule(l){2-8} 
 & \begin{tabular}[c]{@{}c@{}}easy\\ hard\end{tabular} & \begin{tabular}[c]{@{}c@{}}EA\\ SA\end{tabular} & \begin{tabular}[c]{@{}c@{}}SA\\ EA\end{tabular} & \begin{tabular}[c]{@{}c@{}}EA\\ FGA\end{tabular} & \begin{tabular}[c]{@{}c@{}}FGA\\ EA\end{tabular} & \begin{tabular}[c]{@{}c@{}}SA\\ FGA\end{tabular} & \begin{tabular}[c]{@{}c@{}}FGA\\ SA\end{tabular} \\ \midrule
\multirow{4}{*}{\begin{tabular}[c]{@{}c@{}}performance \\ ratio\end{tabular}} & average & 1.001 & 1.000 & 1.022 & 0.980 & 1.021 & 0.980 \\
 & std & 0.008 & 0.008 & 0.010 & 0.010 & 0.012 & 0.011 \\
 & min & 0.954 & 0.973 & 0.992 & 0.939 & 0.995 & 0.930 \\
 & max & 1.029 & 1.049 & 1.067 & 1.009 & 1.076 & 1.006 \\ \midrule
\multirow{4}{*}{\begin{tabular}[c]{@{}c@{}}discounting\\ expected\\ approximation\\ ratio\end{tabular}} & average & 0.955 & 0.954 & 0.968 & 0.928 & 0.966 & 0.928 \\
 & std & 0.020 & 0.017 & 0.018 & 0.022 & 0.017 & 0.025 \\
 & min & 0.837 & 0.858 & 0.876 & 0.840 & 0.890 & 0.827 \\
 & max & 0.999 & 0.993 & 1.011 & 0.979 & 1.018 & 0.981 \\ \bottomrule
\end{tabular}

\end{table}

First, we did some experiments using the graphs 
that are already implemented in IOHProfiler   
which are "frb" graphs with 450, 595, and 760 nodes. We test different combinations of five different baseline algorithms namely (1+1)~EA, fast genetic algorithm (FGA),  simulated annealing (SA), random local search(RLS) and greedy hill climber (GHC) on these instances. The performance ratios that we got for these instances were all very close to 1, which means the performances of all these baseline algorithms are very similar. We could see some ratios larger than 1 when using Equation \ref{eq:FE1} as the fitness function, but these results were not stable and had a very large standard deviation. To test the reliability of these instances, we run the algorithms on them again to see if we get similar performance ratios. The results show these ratios are not reliable and very hard to replicate.
Therefore, we proposed Equation \ref{eq:FE2} as our fitness function to get more stable results. The performance ratios using this formula as the fitness function were much more stable and all the ratios were very close to 1. This means it is not possible to make the baseline algorithms hard for any of these graphs by changing expected values and variances of the costs. This is because these graphs are very dense and in spite of them being relatively large graphs, we can cover a lot of the nodes by just picking a very small subset of the nodes. For example, in the "frb" graph with 450 nodes, we were able to cover around 430 nodes by just picking around 10 nodes in each solution. This makes the problem too easy for the algorithms to solve regardless of the distribution of the nodes' weights.

We proceed to use some new graphs for our experiments which are much more sparse and can show the difference between the performance of baseline algorithms \cite{DBLP:conf/aaai/RossiA15}. These graphs have 204, 379, 425, 500, 615, and 715 nodes with 682, 914, 1300, 2400, 2400, and 3000 edges, respectively. With the setting we currently have for the constraint budget, we are able to cover approximately 70 percent of nodes of these graphs by picking a subset containing around 15 percent of nodes, thus, we do not have the problem that we had with "frb" graphs for our experiments.
First, we generate 1000 random instances using the "netscience" graph and calculated the performance difference between 3 baseline algorithms using performance ratio (Equation \ref{eq:FE1}) and discounting approximation ratio (Equation \ref{eq:FE2}) with 0.99 confidence level, to see the range of performance ratios we can get for these algorithms in a sparse graph. We chose FGA, EA, and SA, with 10,000 and 100,000 function 
evaluation budget, for these experiments. The summary of the results for these instances is represented in Table \ref{tab:random10} and \ref{tab:random100}. By investigating the ratios that we got for these random instances we can determine whether it is feasible to create instances that are hard for an algorithm and easy for the other one. As can be seen in these tables, SA and EA have very similar performance, but we can see a good range in the performance ratio of EA and FGA. Hence, it is possible to get discriminating instances that are easy to solve for (1+1)~EA and hard for FGA. 
Afterward, we generate instances with differentiating constraints using the two proposed fitness functions. The detailed setup and results of these experiments are presented in the following sections.

\begin{table}
\centering
\caption{Performance ratios of 10 evolved instances which are easy to solve for EA and hard for FGA, using approximation ratio (Equation \ref{eq:FE1}) as fitness function} \label{e/f-EA}
\renewcommand{\tabcolsep}{7pt}
\renewcommand{\arraystretch}{1.1}
\begin{tabular}{ccllllll}
\hline
\multicolumn{8}{c}{EA/FGA} \\ \hline
\multirow{2}{*}{\begin{tabular}[c]{@{}c@{}}Graph\end{tabular}} & \multirow{2}{*}{} & \multicolumn{2}{c}{EA-10} & \multicolumn{2}{c}{EA-15} & \multicolumn{2}{c}{EA-20} \\ \cline{3-8} 
 &  & \multicolumn{1}{c}{average} & \multicolumn{1}{c}{std} & \multicolumn{1}{c}{average} & \multicolumn{1}{c}{std} & \multicolumn{1}{c}{average} & \multicolumn{1}{c}{std} \\ \hline

 \multirow{4}{*}{lp-recipe} 
& average & 1.1367 & 0.0937 & 1.1249 & 0.0781 & 1.1249 & 0.0799\\ 
&std & 0.0221 & 0.0220 & 0.0179 & 0.0250 & 0.0188 & 0.0238  \\ 
&min & 1.1028 & 0.0535 & 1.0964 & 0.0517 & 1.0944 & 0.0442  \\ 
& max & 1.1840 & 0.1314 & 1.1585 & 0.1333 & 1.1583 & 0.1403 \\  \hline
       
\multirow{4}{*}{ca-netscience} & average & 1.1872 & 0.0950 & 1.1932 & 0.0997 & 1.1970 & 0.1029 \\
 & std & 0.0161 & 0.0192 & 0.0278 & 0.0317 & 0.0266 & 0.0328 \\
 & min & 1.1549 & 0.0559 & 1.1466 & 0.0580 & 1.1544 & 0.0564 \\
 & max & 1.2113 & 0.1209 & 1.2392 & 0.1672 & 1.2350 & 0.1538 \\ \hline
\multirow{4}{*}{impcol-d} & average & 1.1822 & 0.0938 & 1.1862 & 0.0832 & 1.1825 & 0.0906 \\
 & std & 0.0170 & 0.0215 & 0.0158 & 0.0183 & 0.0200 & 0.0106 \\
 & min & 1.1507 & 0.0623 & 1.1555 & 0.0489 & 1.1381 & 0.0694 \\
 & max & 1.2106 & 0.1395 & 1.2046 & 0.1134 & 1.2140 & 0.1036 \\ \hline

\multirow{4}{*}{random graph}
&average & 1.1345 & 0.0670 & 1.1365 & 0.0668 & 1.1415 & 0.0608  \\ 
&std & 0.0099 & 0.0221 & 0.0100 & 0.0116 & 0.0090 & 0.0150  \\
&min & 1.1193 & 0.0292 & 1.1169 & 0.0457 & 1.1286 & 0.0425  \\ 
&max & 1.1483 & 0.1055 & 1.1498 & 0.0903 & 1.1623 & 0.0950 \\ \hline

 \multirow{4}{*}{lp-agg}
 &average & 1.2282 & 0.1368 & 1.2695 & 0.1633 & 1.2451 & 0.1495  \\
&std & 0.0229 & 0.0273 & 0.0431 & 0.0455 & 0.0383 & 0.0372  \\
&min & 1.1932 & 0.0973 & 1.2002 & 0.0973 & 1.1665 & 0.0834  \\
&max & 1.2711 & 0.1961 & 1.3498 & 0.2486 & 1.2888 & 0.2077 \\
 \hline

 \multirow{4}{*}{can-715} 
 &average & 1.2213 & 0.0947 & 1.2250 & 0.0967 & 1.2371 & 0.1044  \\
&std & 0.0235 & 0.0112 & 0.0236 & 0.0133 & 0.0280 & 0.0259  \\
&min & 1.1798 & 0.0806 & 1.1758 & 0.0759 & 1.1912 & 0.0622  \\
&max & 1.2622 & 0.1188 & 1.2560 & 0.1151 & 1.2916 & 0.1526 \\
 \hline
\end{tabular}
\end{table}

\begin{table}
\centering
\caption{Performance ratios of 10 evolved instances which are easy to solve for EA and hard for GHC, using approximation ratio (Equation \ref{eq:FE1}) as fitness function} \label{e/g-EA}
\renewcommand{\tabcolsep}{7pt}
\renewcommand{\arraystretch}{1.1}
\begin{tabular}{ccllllll}
\hline
\multicolumn{8}{c}{EA/GHC} \\ \hline
\multirow{2}{*}{\begin{tabular}[c]{@{}c@{}}Graph\end{tabular}} & \multirow{2}{*}{} & \multicolumn{2}{c}{EA-10} & \multicolumn{2}{c}{EA-15} & \multicolumn{2}{c}{EA-20} \\ \cline{3-8} 
 &  & \multicolumn{1}{c}{average} & \multicolumn{1}{c}{std} & \multicolumn{1}{c}{average} & \multicolumn{1}{c}{std} & \multicolumn{1}{c}{average} & \multicolumn{1}{c}{std} \\ \hline

 \multirow{4}{*}{lp-recipe} 
&average & 5.4756 & 2.4641 & 5.4372 & 2.4924 & 5.6844 & 2.2155  \\
&std & 0.5938 & 0.9735 & 0.5960 & 0.5276 & 0.8126 & 0.4771  \\ 
&min & 4.7795 & 1.0572 & 4.6967 & 1.4993 & 4.5117 & 1.6567  \\ 
&max & 6.5275 & 4.4418 & 6.2857 & 3.1053 & 7.3508 & 3.3976 \\ \hline
       
\multirow{4}{*}{ca-netscience}
&average & 3.1806 & 1.5766 & 3.5060 & 1.5808 & 3.4475 & 1.3911  \\
&std & 0.5835 & 0.4375 & 0.7493 & 0.3796 & 0.5952 & 0.3287  \\
&min & 2.0304 & 0.7927 & 2.3860 & 1.0251 & 2.6490 & 0.7088  \\
&max & 4.3021 & 2.3327 & 4.7075 & 2.2537 & 4.5424 & 2.0070 \\ \hline
\multirow{4}{*}{impcol-d} 
&average & 3.0914 & 1.2501 & 2.9508 & 1.3684 & 2.7282 & 1.1520  \\
&std & 0.3698 & 0.3272 & 0.3436 & 0.3484 & 0.2536 & 0.2375  \\
&min & 2.4937 & 0.6821 & 2.3052 & 0.9772 & 2.3550 & 0.8851  \\ 
&max & 3.8645 & 1.7602 & 3.5383 & 2.1413 & 3.1420 & 1.6934 \\  \hline

\multirow{4}{*}{random graph}
&average & 2.7540 & 0.3240 & 2.7620 & 0.3032 & 2.5653 & 0.2641  \\
&std & 0.2007 & 0.0922 & 0.1259 & 0.0736 & 0.1310 & 0.0445  \\
&min & 2.4368 & 0.1527 & 2.5445 & 0.2200 & 2.3054 & 0.1589  \\
&max & 3.0902 & 0.4727 & 3.0773 & 0.4156 & 2.8135 & 0.3097 \\ \hline

 \multirow{4}{*}{lp-agg}
 &average & 7.9491 & 1.8209 & 7.8872 & 2.0924 & 7.7486 & 1.7860  \\
&std & 1.3456 & 0.7589 & 1.4564 & 0.6016 & 1.0273 & 0.4704  \\
&min & 5.2590 & 0.8080 & 5.5942 & 0.8791 & 6.2278 & 1.1401  \\
&max & 10.4909 & 3.1013 & 10.1769 & 3.0805 & 9.1818 & 2.4519 \\
 \hline

 \multirow{4}{*}{can-715} 
&average & 2.4215 & 0.3719 & 2.3662 & 0.2416 & 2.5636 & 0.4798  \\
&std & 0.1876 & 0.1015 & 0.2563 & 0.0544 & 0.2995 & 0.2427  \\
&min & 2.1893 & 0.2163 & 1.9597 & 0.1522 & 2.1895 & 0.2067  \\
&max & 2.7909 & 0.5580 & 3.0033 & 0.3056 & 3.0955 & 0.9923 \\
 \hline
\end{tabular}
\end{table}

\subsection{Experimental Setting}

\begin{sidewaystable*}

\centering
\caption{Performance ratios of 10 evolved instances which are easy to solve for EA and hard for FGA, using discounting approximation ratio (Equation \ref{eq:FE2}) as fitness function} \label{e/f_new}

\renewcommand{\arraystretch}{1.3}
\renewcommand{\tabcolsep}{2pt}
\small
\begin{tabular}{@{}ccllllllllllllllllll@{}}
\toprule
 \multicolumn{20}{c}{\textbf{EA/FGA}}\\ \midrule
   \multicolumn{2}{c}{} & \multicolumn{9}{c}{ confidence level = 0.9} &   \multicolumn{9}{c}{ confidence level = 0.99}\\ \cmidrule(l){3-20}

\multirow{2}{*}{Graph} & \multirow{2}{*}{} & \multicolumn{3}{c}{$\text{EA}_\text{D}\text{-10}$} & \multicolumn{3}{c}{$\text{EA}_\text{D}\text{-15}$} & \multicolumn{3}{c}{$\text{EA}_\text{D}\text{-20}$} 
\multirow{2}{*}{} & \multicolumn{3}{c}{$\text{EA}_\text{D}\text{-10}$} & \multicolumn{3}{c}{$\text{EA}_\text{D}\text{-15}$} & \multicolumn{3}{c}{$\text{EA}_\text{D}\text{-20}$} 
\\ \cmidrule(l){3-20} 
 &  & \multicolumn{1}{c}{\begin{tabular}[c]{@{}c@{}}Function\\ value\end{tabular}} & \multicolumn{1}{c}{average} & \multicolumn{1}{c}{std} & \multicolumn{1}{c}{\begin{tabular}[c]{@{}c@{}}Function\\ value\end{tabular}} & \multicolumn{1}{c}{average} & \multicolumn{1}{c}{std} & \multicolumn{1}{c}{\begin{tabular}[c]{@{}c@{}}Function\\ value\end{tabular}} & \multicolumn{1}{c}{average} & \multicolumn{1}{c}{std}

& \multicolumn{1}{c}{\begin{tabular}[c]{@{}c@{}}Function\\ value\end{tabular}} & \multicolumn{1}{c}{average} & \multicolumn{1}{c}{std} & \multicolumn{1}{c}{\begin{tabular}[c]{@{}c@{}}Function\\ value\end{tabular}} & \multicolumn{1}{c}{average} & \multicolumn{1}{c}{std} & \multicolumn{1}{c}{\begin{tabular}[c]{@{}c@{}}Function\\ value\end{tabular}} & \multicolumn{1}{c}{average} & \multicolumn{1}{c}{std}

\\ \midrule

 \multirow{4}{*}{lp-recipe} 
&average & 1.0675 & 1.1093 & 0.0327 & 1.0652 & 1.0963 & 0.0243 & 1.0654 & 1.1063 & 0.0319  & 1.0388 & 1.0869 & 0.0207 & 1.0404 & 1.0845 & 0.0190 & 1.0347 & 1.0835 & 0.0210 \\ 
&std & 0.0151 & 0.0237 & 0.0089 & 0.0128 & 0.0164 & 0.0060 & 0.0169 & 0.0229 & 0.0098 & 0.0119 & 0.0197 & 0.0057 & 0.0103 & 0.0205 & 0.0065 & 0.0060 & 0.0142 & 0.0062 \\ 
&min & 1.0467 & 1.0800 & 0.0237 & 1.0462 & 1.0660 & 0.0155 & 1.0482 & 1.0719 & 0.0128 & 1.0268 & 1.0666 & 0.0135 & 1.0236 & 1.0565 & 0.0083 & 1.0229 & 1.0570 & 0.0107  \\ 
&max & 1.0960 & 1.1598 & 0.0498 & 1.0902 & 1.1227 & 0.0403 & 1.0936 & 1.1453 & 0.0437 & 1.0642 & 1.1306 & 0.0330 & 1.0590 & 1.1194 & 0.0290 & 1.0477 & 1.1078 & 0.0298 \\ 
 \midrule

\multirow{4}{*}{ca-netscience} 
& average & 1.1032 & 1.1639 & 0.0474 & 1.1022 & 1.1582 & 0.0437 & 1.0911 & 1.1474 & 0.0440 & 1.064 & 1.123 & 0.026 & 1.060 & 1.122 & 0.027 & 1.066 & 1.137 & 0.031 \\
 & std & 0.0320 & 0.0437 & 0.0113 & 0.0144 & 0.0234 & 0.0132 & 0.0127 & 0.0212 & 0.0107 & 0.017 & 0.025 & 0.006 & 0.015 & 0.030 & 0.008 & 0.013 & 0.015 & 0.007 \\
 & min & 1.0586 & 1.1012 & 0.0333 & 1.0765 & 1.1072 & 0.0240 & 1.0716 & 1.1108 & 0.0299 & 1.040 & 1.063 & 0.010 & 1.034 & 1.064 & 0.011 & 1.049 & 1.115 & 0.018\\
 & max & 1.1855 & 1.2796 & 0.0734 & 1.1235 & 1.1938 & 0.0671 & 1.1090 & 1.1756 & 0.0620  & 1.100 & 1.167 & 0.033 & 1.082 & 1.150 & 0.037 & 1.088 & 1.168 & 0.042 \\ \midrule

\multirow{4}{*}{impcol-d}
& average & 1.1060 & 1.1522 & 0.0361 & 1.1027 & 1.1456 & 0.0335 & 1.1121 & 1.1541 & 0.0328 & 1.077 & 1.132 & 0.023 & 1.078 & 1.144 & 0.028 & 1.080 & 1.137 & 0.024\\
 & std & 0.0120 & 0.0164 & 0.0067 & 0.0070 & 0.0102 & 0.0081 & 0.0129 & 0.0190 & 0.0081 & 0.009 & 0.016 & 0.005 & 0.015 & 0.035 & 0.011 & 0.011 & 0.017 & 0.006\\
 & min & 1.0842 & 1.1305 & 0.0265 & 1.0910 & 1.1302 & 0.0198 & 1.0998 & 1.1298 & 0.0231& 1.062 & 1.106 & 0.017 & 1.051 & 1.089 & 0.017 & 1.061 & 1.097 & 0.009  \\
 & max & 1.1277 & 1.1834 & 0.0495 & 1.1116 & 1.1614 & 0.0425 & 1.1463 & 1.1920 & 0.0490  & 1.088 & 1.158 & 0.035 & 1.103 & 1.225 & 0.055 & 1.096 & 1.157 & 0.033 \\ 
\midrule

 \multirow{4}{*}{random graph} 
&average & 1.0790 & 1.1160 & 0.0288 & 1.0881 & 1.1193 & 0.0243 & 1.0856 & 1.1142 & 0.0223 & 1.0591 & 1.1045 & 0.0195 & 1.0616 & 1.0998 & 0.0165 & 1.0556 & 1.0900 & 0.0148  \\ 
&std & 0.0050 & 0.0145 & 0.0091 & 0.0118 & 0.0125 & 0.0053 & 0.0101 & 0.0106 & 0.0053 & 0.0106 & 0.0187 & 0.0044 & 0.0066 & 0.0148 & 0.0051 & 0.0051 & 0.0121 & 0.0043  \\ 
&min & 1.0733 & 1.0925 & 0.0150 & 1.0613 & 1.0908 & 0.0169 & 1.0744 & 1.0955 & 0.0142 & 1.0458 & 1.0735 & 0.0108 & 1.0526 & 1.0771 & 0.0102 & 1.0469 & 1.0702 & 0.0083  \\ 
&max & 1.0909 & 1.1353 & 0.0421 & 1.1041 & 1.1370 & 0.0370 & 1.1110 & 1.1298 & 0.0290 & 1.0807 & 1.1357 & 0.0267 & 1.0735 & 1.1212 & 0.0256 & 1.0645 & 1.1067 & 0.0216\\
 \midrule

 \multirow{4}{*}{lp-agg} 
&average & 1.1077 & 1.1673 & 0.0465 & 1.1194 & 1.1810 & 0.0481 & 1.1211 & 1.1914 & 0.0548  & 1.0660 & 1.1510 & 0.0365 & 1.0764 & 1.1813 & 0.0451 & 1.0660 & 1.1510 & 0.0365 \\ 
&std & 0.0106 & 0.0162 & 0.0105 & 0.0228 & 0.0419 & 0.0192 & 0.0204 & 0.0301 & 0.0155 & 0.0247 & 0.0533 & 0.0160 & 0.0226 & 0.0413 & 0.0142 & 0.0247 & 0.0533 & 0.0160 \\ 
&min & 1.0892 & 1.1451 & 0.0280 & 1.0800 & 1.1165 & 0.0212 & 1.0953 & 1.1415 & 0.0346  & 1.0390 & 1.0804 & 0.0162 & 1.0454 & 1.1233 & 0.0284 & 1.0390 & 1.0804 & 0.0162  \\ 
&max & 1.1241 & 1.1976 & 0.0622 & 1.1631 & 1.2657 & 0.0800 & 1.1558 & 1.2446 & 0.0814 & 1.1201 & 1.2489 & 0.0668 & 1.1149 & 1.2426 & 0.0672 & 1.1201 & 1.2489 & 0.0668 \\
 \midrule

 \multirow{4}{*}{can-715} 
&average & 1.1302 & 1.1902 & 0.0468 & 1.1349 & 1.1931 & 0.0454 & 1.1331 & 1.1940 & 0.0475  & 1.0946 & 1.1687 & 0.0319 & 1.1003 & 1.1742 & 0.0318 & 1.0904 & 1.1511 & 0.0261\\ 
&std & 0.0126 & 0.0220 & 0.0163 & 0.0171 & 0.0331 & 0.0172 & 0.0207 & 0.0185 & 0.0065 & 0.0144 & 0.0286 & 0.0094 & 0.0158 & 0.0132 & 0.0083 & 0.0195 & 0.0313 & 0.0066   \\ 
&min & 1.1058 & 1.1540 & 0.0186 & 1.1096 & 1.1524 & 0.0228 & 1.1085 & 1.1666 & 0.0372   & 1.0674 & 1.1313 & 0.0203 & 1.0758 & 1.1510 & 0.0197 & 1.0716 & 1.1271 & 0.0169\\ 
&max & 1.1560 & 1.2294 & 0.0667 & 1.1638 & 1.2456 & 0.0663 & 1.1741 & 1.2253 & 0.0565& 1.1206 & 1.2169 & 0.0493 & 1.1294 & 1.1994 & 0.0490 & 1.1382 & 1.2344 & 0.0414  \\

 \bottomrule
\end{tabular}
\end{sidewaystable*}
For our experiments, we use six different graphs. Five of those graphs are obtained from \cite{DBLP:conf/aaai/RossiA15} namely: "lp-recipe" with 204 nodes and 682 edges, "ca-netscience" with 379 nodes and 914 edges, "impcol-d" with 425 nodes and 13k edges, "lp-agg" with 615 nodes and 2.4K edges, and "can-715" with 715 nodes and 3k edges. We also generate an undirected random graph with 500 nodes where each edge is present with probability $p = 0.02$. We refer to this graph as "random graph" in the tables.

 To initiate our (1+1)~EA for evolving instances, we generate an instance in which the expected values ($\mu_i$) and variances ($\sigma^2_i$) associated with costs of each vertex $i$, are chosen uniformly at random respectively from $(0,1000)$ and $(0,\;\mu_i^2/3)$. To calculate the suitable budget for each instance, we use $B = n/30 * 500$, where n is the number of nodes in graphs. We set the probability of constraint violation as $\alpha =0.05$.
 To evaluate the performance of each algorithm we run each of them for 10 independent runs with 10,000 function evaluations. We set the value of $P(A)$ to $\epsilon=10^{-2}$ for the runs that do not reach a feasible solution. At the end of each run, we will get one discriminating instance which results in $10$ instances for each graph and pair of algorithms.
 
For the mutation, the expected value and variance of each node change with the probability of $Pm=1$. We choose $a$ and $b$ from normal distributions with expected values of 0 and variances of $\sigma_1$ and $\sigma_2$, respectively. We compare 3 different settings for the value of $\sigma_1$ and $\sigma_2$ in mutation function: ($\sigma_1=10, \sigma_2=33$), ($\sigma_1=15, \sigma_2=75$), and ($\sigma_1=20, \sigma_2=133$).
These scales are adapted using a 1/5 success rule update. After the mutation process is completed for the instance, the value of the budget will be updated accordingly. Lastly, we pass the individual with better fitness value to the next generation. In our experimental results, we compare three different fitness functions. One fitness function does not consider the standard deviation by using Equation \ref{eq:FE1} and the other two fitness functions use Equation \ref{eq:FE2} with confidence levels 0.9 and 0.99.

All the experiments are done for 10 independent runs and 10,000 function evaluations for each graph and pair of algorithms.

\subsection{Evolving Discriminating Instances based on approximation ratio}

 In this section, we use the (1+1)~EA to generate easy/hard instances, using the traditional approximation ratio (Equation 5) as our fitness function. These instances are easy to solve by EA and hard to solve by FGA and GHC with a 10,000 function evaluation budget, using 3 different settings for mutation. We change the mutation function by changing the values of standard deviation $\sigma_1$ and $\sigma_2$ in the normal distribution. By increasing the sigma value, we are making more drastic changes to create offspring, which helps with better exploration of solution space. These values are as follows: ($\sigma_1=10, \sigma_2=33$), ($\sigma_1=15, \sigma_2=75$), and ($\sigma_1=20, \sigma_2=133$).
To evolve these discriminating instances we run the (1+1)~EA (Algorithm \ref{alg:main}), for 10,000 function evaluations and 10 independent runs, and calculate the average of these runs as performance ratio. The results of these experiments are presented in Table \ref{e/f-EA} and Table \ref{e/g-EA}. Where EA-10, EA-15, and EA-20 mean we are using these values as the value of $\sigma_1$ and their aforementioned corresponding values of $\sigma_2$ in the mutation function.

\begin{sidewaystable*}
\centering
\caption{Performance ratios of 10 evolved instances which are easy to solve for EA and hard for GHC, using discounting approximation ratio (Equation \ref{eq:FE2}) as fitness function} \label{e/g_new}
\renewcommand{\arraystretch}{1.3}
\renewcommand{\tabcolsep}{2pt}
\small
\begin{tabular}{@{}ccllllllllllllllllll@{}}
\toprule
 \multicolumn{20}{c}{EA/GHC}\\ \midrule
   \multicolumn{2}{c}{} & \multicolumn{9}{c}{ confidence level = 0.9} &   \multicolumn{9}{c}{ confidence level = 0.99}\\ \cmidrule(l){3-20}

\multirow{2}{*}{Graph} & \multirow{2}{*}{} & \multicolumn{3}{c}{$\text{EA}_\text{D}\text{-10}$} & \multicolumn{3}{c}{$\text{EA}_\text{D}\text{-15}$} & \multicolumn{3}{c}{$\text{EA}_\text{D}\text{-20}$} 
\multirow{2}{*}{} & \multicolumn{3}{c}{$\text{EA}_\text{D}\text{-10}$} & \multicolumn{3}{c}{$\text{EA}_\text{D}\text{-15}$} & \multicolumn{3}{c}{$\text{EA}_\text{D}\text{-20}$} 
\\ \cmidrule(l){3-20} 
 &  & \multicolumn{1}{c}{\begin{tabular}[c]{@{}c@{}}Function\\ value\end{tabular}} & \multicolumn{1}{c}{average} & \multicolumn{1}{c}{std} & \multicolumn{1}{c}{\begin{tabular}[c]{@{}c@{}}Function\\ value\end{tabular}} & \multicolumn{1}{c}{average} & \multicolumn{1}{c}{std} & \multicolumn{1}{c}{\begin{tabular}[c]{@{}c@{}}Function\\ value\end{tabular}} & \multicolumn{1}{c}{average} & \multicolumn{1}{c}{std}

& \multicolumn{1}{c}{\begin{tabular}[c]{@{}c@{}}Function\\ value\end{tabular}} & \multicolumn{1}{c}{average} & \multicolumn{1}{c}{std} & \multicolumn{1}{c}{\begin{tabular}[c]{@{}c@{}}Function\\ value\end{tabular}} & \multicolumn{1}{c}{average} & \multicolumn{1}{c}{std} & \multicolumn{1}{c}{\begin{tabular}[c]{@{}c@{}}Function\\ value\end{tabular}} & \multicolumn{1}{c}{average} & \multicolumn{1}{c}{std}
\\ \midrule

 \multirow{4}{*}{lp-recipe} 
&average & 3.4168 & 3.9845 & 0.4430 & 3.8135 & 4.4851 & 0.5241 & 3.5307 & 4.2904 & 0.5928  & 3.0131 & 3.7594 & 0.3208 & 3.0589 & 3.5972 & 0.2314 & 3.0328 & 3.6474 & 0.2642 \\ &std & 0.5439 & 0.5341 & 0.0636 & 0.6456 & 0.6661 & 0.1610 & 0.5791 & 0.9315 & 0.3289 & 0.4543 & 0.7400 & 0.1691 & 0.3382 & 0.4199 & 0.0757 & 0.6064 & 0.7944 & 0.0862 \\
&min & 2.6471 & 3.2361 & 0.3458 & 2.8755 & 3.1347 & 0.2023 & 2.8076 & 3.1801 & 0.2907  & 2.2386 & 2.5728 & 0.1298 & 2.5042 & 2.8256 & 0.1382 & 1.6210 & 1.8180 & 0.0847  \\ 
&max & 4.4777 & 5.0407 & 0.5755 & 4.9337 & 5.4316 & 0.7460 & 4.8369 & 6.7191 & 1.4687 & 3.8045 & 5.1810 & 0.5921 & 3.7582 & 4.1841 & 0.3670 & 3.8299 & 4.6328 & 0.3452 \\
 \midrule

\multirow{4}{*}{ca-netscience} 
&average & 2.3707 & 2.7382 & 0.2868 & 2.1653 & 2.5822 & 0.3253 & 1.9326 & 2.2911 & 0.2798  & 1.5514 & 1.9866 & 0.1871 & 1.8854 & 2.3348 & 0.1932 & 1.5097 & 1.7516 & 0.1040 \\
&std & 0.5819 & 0.7148 & 0.1686 & 0.5891 & 0.8152 & 0.2174 & 0.4325 & 0.6249 & 0.1704 & 0.2221 & 0.3990 & 0.1258 & 0.4997 & 0.7103 & 0.1118 & 0.1452 & 0.3013 & 0.0837   \\
&min & 1.5779 & 1.8447 & 0.1251 & 1.5267 & 1.6441 & 0.0916 & 1.3055 & 1.4585 & 0.0952  & 1.2006 & 1.2635 & 0.0270 & 1.3847 & 1.6071 & 0.0867 & 1.2327 & 1.2974 & 0.0278 \\
&max & 3.3333 & 3.7626 & 0.7134 & 3.5344 & 4.3615 & 0.6687 & 2.8684 & 3.6722 & 0.6272& 1.9342 & 2.6534 & 0.3739 & 3.0732 & 4.0209 & 0.4074 & 1.7274 & 2.3654 & 0.3202 \\

\midrule
\multirow{4}{*}{impcol-d}
&average & 2.1158 & 2.3657 & 0.1950 & 1.8873 & 2.0661 & 0.1395 & 2.0044 & 2.2205 & 0.1686   & 1.6832 & 1.8979 & 0.0923 & 1.8740 & 2.2593 & 0.1656 & 1.8767 & 2.1691 & 0.1257  \\ &std & 0.3501 & 0.3931 & 0.0493 & 0.1637 & 0.1776 & 0.0349 & 0.2660 & 0.3088 & 0.0596 & 0.0967 & 0.1414 & 0.0382 & 0.1892 & 0.2846 & 0.0893 & 0.1013 & 0.1663 & 0.0400  \\ 
&min & 1.7219 & 1.9260 & 0.1283 & 1.6149 & 1.7755 & 0.0841 & 1.6025 & 1.8706 & 0.0802 & 1.5110 & 1.6859 & 0.0332 & 1.6799 & 1.8812 & 0.0661 & 1.7085 & 1.9715 & 0.0770  \\ 
&max & 3.0544 & 3.4339 & 0.2962 & 2.1329 & 2.3166 & 0.2074 & 2.4918 & 2.7352 & 0.2643 & 1.8398 & 2.1256 & 0.1495 & 2.3135 & 2.8425 & 0.3855 & 2.1024 & 2.5337 & 0.2127\\
\midrule

 \multirow{4}{*}{random graph} 
&average & 2.5276 & 2.6674 & 0.1091 & 2.5538 & 2.7515 & 0.1543 & 2.5013 & 2.6889 & 0.1464 & 2.3641 & 2.5974 & 0.1003 & 2.2347 & 2.4757 & 0.1036 & 2.3418 & 2.5736 & 0.0996  \\ &std & 0.1918 & 0.1898 & 0.0278 & 0.1371 & 0.1768 & 0.0423 & 0.1717 & 0.1720 & 0.0452   & 0.1378 & 0.1522 & 0.0213 & 0.1371 & 0.1686 & 0.0296 & 0.1545 & 0.1931 & 0.0338\\ 
&min & 2.1784 & 2.2949 & 0.0554 & 2.2694 & 2.4070 & 0.1074 & 2.3235 & 2.5173 & 0.0768  & 2.1145 & 2.3481 & 0.0748 & 2.0207 & 2.1457 & 0.0538 & 2.1034 & 2.2638 & 0.0419 \\ 
&max & 2.8396 & 2.9769 & 0.1633 & 2.7733 & 3.1031 & 0.2573 & 2.8663 & 3.0201 & 0.2284  & 2.5552 & 2.8202 & 0.1416 & 2.5070 & 2.7596 & 0.1622 & 2.6601 & 2.8079 & 0.1438\\ 
 \midrule

 \multirow{4}{*}{lp-agg} 
&average & 4.6929 & 6.4507 & 1.3717 & 6.0234 & 7.7465 & 1.3446 & 7.0113 & 8.2811 & 0.9908 & 5.2013 & 7.1553 & 0.8399 & 4.3990 & 6.3712 & 0.8478 & 4.0287 & 6.1497 & 0.9118 \\ &std & 1.7401 & 1.8682 & 0.4611 & 1.9692 & 1.7627 & 0.4969 & 1.8412 & 1.4407 & 0.3763 & 2.0065 & 2.0350 & 0.3807 & 1.9986 & 2.0917 & 0.4689 & 2.1880 & 2.3289 & 0.5404 \\ 
&min & 2.1058 & 2.6361 & 0.4138 & 2.8005 & 4.2175 & 0.6068 & 3.5673 & 5.6600 & 0.5209  & 2.6099 & 2.9550 & 0.1484 & 2.1893 & 2.7582 & 0.1676 & 2.3119 & 2.9119 & 0.2334   \\ 
&max & 7.6195 & 9.4360 & 2.0082 & 9.4929 & 10.8884 & 2.0345 & 9.6697 & 10.3908 & 1.6329  & 7.9825 & 9.8402 & 1.5806 & 8.0128 & 8.7058 & 1.5315 & 8.6851 & 10.0113 & 1.8735 \\ 
 \midrule

 \multirow{4}{*}{can-715} 
&average & 2.0794 & 2.2399 & 0.1252 & 2.1502 & 2.3679 & 0.1698 & 2.0077 & 2.1796 & 0.1342 & 1.9132 & 2.1068 & 0.0832 & 1.8734 & 2.0939 & 0.0948 & 1.8632 & 2.0210 & 0.0678   \\ 
&std & 0.1707 & 0.2080 & 0.0463 & 0.1522 & 0.2364 & 0.0817 & 0.1416 & 0.1897 & 0.0499  & 0.1251 & 0.1788 & 0.0432 & 0.1041 & 0.1546 & 0.0316 & 0.0940 & 0.1407 & 0.0319 \\
&min & 1.8969 & 1.9832 & 0.0674 & 1.9128 & 2.0519 & 0.0774 & 1.8449 & 1.9415 & 0.0401   & 1.7105 & 1.9709 & 0.0306 & 1.7269 & 1.8393 & 0.0381 & 1.7483 & 1.8248 & 0.0325 \\
&max & 2.5398 & 2.7776 & 0.2145 & 2.3728 & 2.7060 & 0.3292 & 2.2524 & 2.5151 & 0.2050  & 2.1277 & 2.5577 & 0.1848 & 2.0700 & 2.3360 & 0.1559 & 2.0150 & 2.3049 & 0.1246 \\ 

 \bottomrule
\end{tabular}
\end{sidewaystable*}

As represented in the tables, we can get a good difference between the performance of the two algorithms by using just this approximation ratio as the fitness function, which is the measure mostly used by previous studies. However, our results show these ratios have very high standard deviations for chance-constrained maximum coverage problem instances. This means using this fitness function does not result in reliable instances in the chance-constrained setting. Especially the standard deviations for instances that are easy to solve by EA and hard to solve by GHC are very high (see Table 4) which increases the risk of getting high ratios by chance and makes replicating the same ratios for these instances almost impossible. For these instances, we get slightly lower standard deviations for all the graphs except "can-715" by using $\sigma_1 = 20$ and $\sigma_2 = 133$ in the mutation function. But this improvement is not enough to give us more reliable instances. To handle the effect of uncertainty in the chance-constrained setting, we use the discounting approximation ratio as the fitness function of the (1+1)~EA to evolve discriminating instances. The experimental investigation for instances generated by using this method is represented in the next section.

\subsection{Evolving Discriminating Instances Using discounting fitness function}
 In this section, we evolve discriminating instances with high confidence using the discounting method. We use Equation \ref{eq:FE2} with two different confidence levels of 0.90 and 0.99 to see the effect of confidence level on the reliability of performance ratios in generated instances. We use the same three settings as before for the mutation function and run the (1+1)~EA with this new fitness function for 10,000 function evaluations and 10 independent runs. The average and standard deviations of the performance ratios of these 10 runs are used to calculate the discounting approximation ratio in the fitness function. The results of these experiments are presented in Table \ref{e/f_new} and \ref{e/g_new}.  Where EA/FGA and EA/GHC show instances that are easy to solve by EA and respectively hard to solve by FGA and GHC, and $\text{EA}_\text{D}\text{-10}$, $\text{EA}_\text{D}\text{-15}$, $\text{EA}_\text{D}\text{-20}$ show the settings that are used in mutation. The function value, average, and standard deviation of the discounting approximation ratios in each setting are reported in these tables. Function value demonstrates the value of discounting approximation ratio,  average shows the performance ratio before reducing the effect of standard deviations for the same instance, and std represents the standard deviation of performance ratios.
 After using this method the standard deviation significantly decreases in all the graphs and algorithms, which shows outstanding improvement in the reliability of the generated instances. This improvement is more visible in the instances that are easy to solve by EA and hard to solve by GHC, since the performance ratios tend to be noisier for these instances. 
 
As expected, using the discounting method leads to slightly lower performance differences for the chosen algorithms. However, the benefit we get from using this method in decreasing the variance of performance ratios for our instances is very significant. As illustrated in the experiments, the effect of discounting uncertainty can be controlled by tuning the confidence level. For instance, using a 0.99 confidence level resulted in lower performance ratios compared to a 0.90 confidence level, because it concentrates more on variances. Having these sets of discriminating benchmark instances with the high level of confidence is very important for reliable future research in the field of algorithm selection.


\section{Conclusions}
In this paper, we addressed the problem of evolving benchmark instances for the chance-constrained maximum coverage problem that are easy for one algorithm and hard for the other, with high confidence. To address this, we proposed a new approach for calculating the fitness function of (1+1)~EA. This new fitness function takes into account the standard deviation of performance ratios, which has not been done before in evolving instances. Using this fitness function limits the influence of uncertainty, resulting in increased stability for the instances. We did experiments using two different fitness functions to measure the performance ratio. One uses the approximation ratio which is the method that has been traditionally used in evolving differentiating instances, and the other is the new discounting approximation ratio which guarantees to increase the confidence level in performance ratios. Our experiments show that we can get a significant difference between the performance of algorithms, while maintaining high reliability, by using this method. The chance-constrained maximum coverage problem instances generated by our approach are much more stable than the approach used by previous studies. These discriminating instances can be used as a comprehensive benchmark set for automated algorithm selection, feature-based algorithm selection and classification and comparison of algorithms for solving chance-constrained problems.

\section*{Acknowledgments}
    This work has been supported by the Australian Research Council through grants DP190103894 and FT200100536.

\bibliographystyle{ACM-Reference-Format}
\bibliography{arxiv}

\appendix

\end{document}